\documentclass{article}

\usepackage[final,nonatbib]{nips_2017}
\usepackage[utf8]{inputenc} 
\usepackage[T1]{fontenc}    
\usepackage{hyperref}       
\usepackage{url}            
\usepackage{booktabs}       
\usepackage{amsfonts}       
\usepackage{nicefrac}       
\usepackage{microtype}      
\usepackage{graphicx}
\usepackage{amsmath}
\usepackage{algorithm}
\usepackage{algpseudocode}
\usepackage{booktabs}
\usepackage{multirow}
\usepackage{amssymb}
\title{Black-Box Attacks against RNN based Malware Detection Algorithms}

\author{Weiwei Hu and Ying Tan\thanks{Prof. Ying Tan is the corresponding author.}\\
Key Laboratory of Machine Perception (MOE), and Department of Machine Intelligence\\
School of Electronics Engineering and Computer Science, Peking University, Beijing, 100871 China\\
\{weiwei.hu, ytan\}@pku.edu.cn}

\begin{document}

\maketitle

\begin{abstract}
Recent researches have shown that machine learning based malware detection algorithms are very vulnerable under the attacks of adversarial examples. These works mainly focused on the detection algorithms which use features with fixed dimension, while some researchers have begun to use recurrent neural networks (RNN) to detect malware based on sequential API features.
This paper proposes a novel algorithm to generate sequential adversarial examples, which are used to attack a RNN based malware detection system.
It is usually hard for malicious attackers to know the exact structures and weights of the victim RNN. A substitute RNN is trained to approximate the victim RNN.
Then we propose a generative RNN to output sequential adversarial examples from the original sequential malware inputs.
Experimental results showed that RNN based malware detection algorithms fail to detect most of the generated malicious adversarial examples, which means the proposed model is able to effectively bypass the detection algorithms.
\end{abstract}

\section{Introduction}
Machine learning has been widely used in various commercial and non-commercial products, and has brought great convenience and profits to human beings. However, recent researches on adversarial examples show that many machine learning algorithms are not robust at all when someone want to crack them on purpose \cite{szegedy2013intriguing,goodfellow2014explaining}. Adding some small perturbations to original samples will make a classifier unable to classify them correctly.

In some security related applications, attackers will try their best to attack any defensive systems to spread their malicious products such as malware. 
Existing machine learning based malware detection algorithms mainly represent programs as feature vectors with fixed dimension and classify them between benign programs and malware \cite{kolter2006learning}. For example, a binary feature vector can be constructed according to the presences or absences of system APIs (i.e. application programming interfaces) in a program \cite{schultz2001data}. Grosse et al. \cite{grosse2016adversarial} and Hu et al. \cite{hu2017generating} have shown that fixed dimensional feature based malware detection algorithms are very vulnerable under the attack of adversarial examples.

Recently, as recurrent neural networks (RNN) become popular, some researchers have tried to use RNN for malware detection and classification \cite{pascanu2015malware,tobiyama2016malware,kolosnjaji2016deep}. The API sequence invoked by a program is used as the input of RNN. RNN will predict whether the program is benign or malware.


This paper tries to validate the security of a RNN based malware detection model when it is attacked by adversarial examples. We proposed a novel algorithm to generate sequential adversarial examples.

Existing researches on adversarial samples mainly focus on images. Images are represented as matrices with fixed dimensions, and the values of the matrices are continuous. API sequences consist of discrete symbols with variable lengths. Therefore, generating adversarial examples for API sequences will become quite different from generating adversarial examples for images.

To generate adversarial examples from API sequences we only consider to insert some irreverent APIs into the original sequences. Removing an API from the API sequence may make the program unable to work. How to insert irreverent APIs into the sequence will be the key to generate adversarial examples.

We propose a generative RNN based approach to generate irreverent APIs and insert them into the original API sequences. A substitute RNN is trained to fit the victim RNN. Gumbel-Softmax \cite{jang2016categorical} is used to smooth the API symbols and deliver gradient information between the generative RNN and the substitute RNN.

\section{Adversarial Examples}
Adversarial examples are usually generated by adding some perturbations to the original samples. Szegedy et al. used a box-constrained L-BFGS to search for an appropriate perturbation which can make a neural network misclassify an image \cite{szegedy2013intriguing}. They found that adversarial examples are able to transfer among different neural networks. Goodfellow et al. proposed the ``fast gradient sign method'' where added perturbations are determined by the gradients of the cost function with respect to inputs \cite{goodfellow2014explaining}. An iterative algorithm to generate adversarial examples was proposed by Papernot et al. \cite{papernot2016limitations}. At each iteration the algorithm only modifies one pixel or two pixels of the image.

Grosse et al. used the iterative algorithm proposed by Papernot et al. \cite{papernot2016limitations} to add some adversarial perturbations to Android malware on about 545 thousand binary features \cite{grosse2016adversarial}. For the best three malware detection models used in their experiments, about 60\% to 70\% malware will become undetected after their adversarial attacks.

Previous algorithms to generate adversarial examples mainly focused on attacking feed-forward neural networks. Papernot et al. migrated these algorithms to attack RNN \cite{papernot2016crafting}. RNN is unrolled along time and existing algorithms for feed-forward neural networks are used to generate adversarial examples for RNN. The limitation of their algorithm is that the perturbations are not truly sequential. For examples, if they want to generate adversarial examples from sentences, they can only replace existing words with others words, but cannot insert words to the original sentences or delete words form the original sentences.

Sometimes it is hard for the attackers to know the structures and parameters of the victim machine learning models. For example, many machine learning models are deployed in remote servers or compiled into binary executables. To attack a black-box victim neural network, Papernot et al. first got the outputs from the victim neural network on their training data, and then trained a substitute neural network to fit the victim neural network \cite{papernot2016practical}. Adversarial examples are generated from the substitute neural network. They also showed that other kinds of machine learning models such as decision trees can also be attacked by using the substitute network to fit them \cite{papernot2016transferability}.

Besides substitute network based approaches, several direct algorithms for black-box attacks have been proposed recently. Narodytska et al. adopted a greedy local search to find a small set of pixels by observing the probability outputs of the victim network after applying perturbations \cite{narodytska2016simple}. Liu et al. used an ensemble-based algorithm to generate adversarial examples and the adversarial examples are able to attack other black-box models due to the transferability of adversarial examples \cite{liu2016delving} .

Several defensive algorithms against adversarial examples have been proposed, such as feature selection \cite{zhang2016adversarial}, defensive distillation \cite{papernot2016distillation} and retraining \cite{li2016general}. However, it is found that the effectiveness of these defensive algorithms is limited, especially under repeated attacks \cite{grosse2016adversarial,chen2016evaluation,carlini2016defensive}.

\section{RNN for Malware Detection}
In this section we will show how to use RNN to detect malware. Malware detection is regarded as a sequential classification problem \cite{pascanu2015malware,tobiyama2016malware,kolosnjaji2016deep}. RNN is used to classify whether an API sequence comes from a benign program or malware.

We will also introduce some variants of RNN in this section. Malware detection model is usually a black box to malware authors, and they need to take the potential variants into consideration when developing attacking algorithms.

Each API is represented as a one-hot vector. Assuming there are $M$ APIs in total, these APIs are numbered from 0 to $M-1$. The feature vector $\boldsymbol{x}$ of an API is an $M$-dimensional binary vector. If the API's number is $i$, the $i$-th dimension of $\boldsymbol{x}$ is 1, and other dimensions are all zeros.

An API sequence is represented as $\boldsymbol{x}_1,\boldsymbol{x}_2,...,\boldsymbol{x}_T$, where $T$ is the length of the sequence. After feeding the input to RNN, the hidden states of RNN can be represented as $\boldsymbol{h}_1,\boldsymbol{h}_2,...,\boldsymbol{h}_T$.

In the basic version of RNN, the hidden state of the last time step $\boldsymbol{h}_{T}$ is used as the representation of the API sequence. The output layer uses $\boldsymbol{h}_{T}$ to compute the probability distribution over the two classes. Then cross entropy is used as the loss function of API sequence classification.

%
%

The first variant of the RNN model introduced here is average pooling \cite{boureau2010theoretical}, which uses the average states across $\boldsymbol{h}_1$ to $\boldsymbol{h}_T$ as the representation of the sequence, instead of the last state $\boldsymbol{h}_{T}$ .

Attention mechanism \cite{bahdanau2014neural} is another variant, which uses weighted average of the hidden states to represent the sequence. Attention mechanism is inspired by the selective nature of human perception. For example, when faced with a picture human beings will focus on some meaningful objects in it, rather than every details of it. Attention mechanism in deep learning makes the model to focus on meaningful parts of inputs. It has shown to be very useful in machine translation \cite{bahdanau2014neural} and image caption \cite{xu2015show}.

An attention function $A$ is defined to map the hidden state to a scalar value, which indicates the importance of the corresponding time step. The attention function is usually a feed-forward neural network. The attention function values across the whole sequence are then normalized according to the formula $\alpha_t={{\exp (A(\boldsymbol{h}_t)))}}/{{\sum\limits_{s = 1}^T {\exp (A(\boldsymbol{h}_s))} }}$, where $\alpha_t$ is the final weight of time step $t$. 


The above RNN models only process the sequence in the forward direction, while some sequential patterns may lie in the backward direction. Bidirectional RNN tries to learn patterns from both directions \cite{schuster1997bidirectional}. In bidirectional RNN, an additional backward RNN is used to process the reversed sequence, i.e. from $\boldsymbol{x}_T$ to $\boldsymbol{x}_1$. The concatenation of the hidden states from both directions is used to calculate the output probability.

\section{Attacking RNN based Malware Detection Algorithms}
Papernot et al. \cite{papernot2016crafting} migrated the adversarial example generation algorithms for feed-forward neural networks to attack RNN by unrolling RNN along time and regarding it as a special kind of feed-forward neural network. However, such model can only replace existing elements in the sequence with other elements, since the perturbations are not truly sequential. This algorithm cannot insert irreverent APIs to the original sequences. The main contribution of this paper is that we proposed a generative RNN based approach to generate sequential adversarial examples, which is able to effectively mine the vulnerabilities in the sequential patterns.

\begin{figure*}[htp]
    \begin{center}
    \graphicspath{{img/}}
    \includegraphics[width = 5.5in]{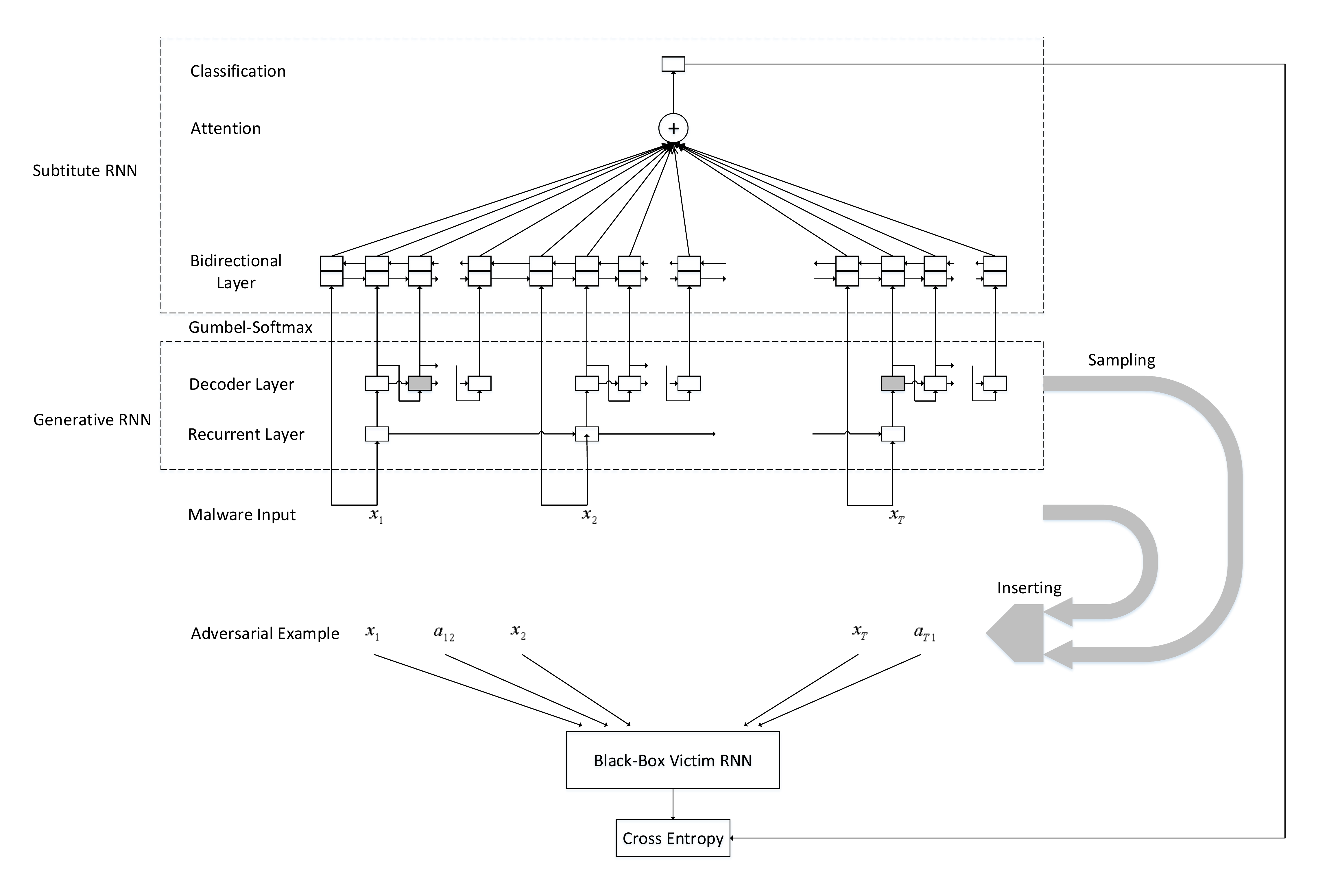}
    \caption{The architecture of the proposed model when trained on malware.}
    \label{fig:gan}
    \end{center}
\end{figure*}

The proposed algorithm consists of a generative RNN and a substitute RNN, as shown in Figure \ref{fig:gan} and Figure \ref{fig:benign}. The generative model is based on a modified version of the sequence to sequence model \cite{sutskever2014sequence}, which takes malware's API sequence as input and generates an adversarial API sequence. The substitute RNN is trained on benign sequences and the Gumbel-Softmax \cite{jang2016categorical} outputs of the generative RNN, in order to fit the black-box victim RNN. The Gumbel-Softmax enables the gradient to back propagate from the substitute RNN to the generative RNN.

\subsection{The Generative RNN}
The input of the generative RNN is a malware API sequence, and the output is the generated sequential adversarial example for the input malware. The generative RNN generates a small piece of API sequence after each API and tries to insert the sequence piece after the API.

For the input sequence $\boldsymbol{x}_1,\boldsymbol{x}_2,...,\boldsymbol{x}_T$, the hidden states of the recurrent layer are $\boldsymbol{h}_1,\boldsymbol{h}_2,...,\boldsymbol{h}_T$. At time step $t$, a small sequence of Gumbel-Softmax output $\boldsymbol{g}_{t1},\boldsymbol{g}_{t2},...,\boldsymbol{g}_{tL}$ with length $L$ is generated based on $\boldsymbol{h}_t$, where $L$ is a hyper-parameter.

Sequence decoder \cite{cho2014learning} is used to generate the small sequence. The decoder RNN uses the formula $\boldsymbol{h}^D_{\tau}=Dec(\boldsymbol{x}^D_\tau,\boldsymbol{h}^D_{\tau-1})$ to update hidden states, where $\boldsymbol{x}^D_\tau$ is the input and $\boldsymbol{h}^D_{\tau}$ is the hidden state of the decoder RNN which is initialized with zero.

Formula \ref{equ:generate1} is used to get the hidden state when generating $\boldsymbol{g}_{t1}$.

\begin{equation}
\label{equ:generate1}
\boldsymbol{h}^D_{1}=Dec(\boldsymbol{h}_t,\boldsymbol{h}^D_{0}=\boldsymbol{0}) .
\end{equation}

When generating the first element at time step $t$, the input is the hidden state $\boldsymbol{h}_t$.

Then a softmax layer is followed to generate the API. Besides the $M$ APIs, we introduce a special null API into the API set. If the null API is generated at time step $\tau$, no API will be inserted to the original sequence at that moment. If we do not use the null API, too many generated APIs will be inserted into the sequence and the resulting sequence will become too long. Allowing null API will make the final sequence shorter. Since the $M$ valid APIs have been numbered from 0 to $M-1$, the null API is numbered as $M$.

The softmax layer will have $M+1$ elements, which is calculated as $\boldsymbol{\pi}_{t1}=softmax(\boldsymbol{W}_{s}\boldsymbol{h}^D_{1})$, where $\boldsymbol{W}_{s}$ is the weights to map the hidden state to the output layer.

%

Then we can sample an API from $\boldsymbol{\pi}_{t1}$. Let the one-hot representation of the sampled API be $\boldsymbol{a}_{t1}$.

The sampled API is a discrete symbol. If we give $\boldsymbol{a}_{t1}$ to the substitute RNN, we are unable to get the gradients from the substitute RNN and thus unable to train the generative RNN.

Gumbel-Softmax is recently proposed to approximate one-hot vectors with differentiable representations \cite{jang2016categorical}. The Gumbel-Softmax output $\boldsymbol{g}_{t1}$ has the same dimension with $\boldsymbol{\pi}_{t1}$. The $i$-th element of $\boldsymbol{g}_{t1}$ is calculated by Formula \ref{equ:generate-Gumbel-softmax}.

\begin{equation}
\label{equ:generate-Gumbel-softmax}
\boldsymbol{g}_{t1}^i = \frac{\exp((\log(\boldsymbol{\pi}^i_{t1})+z_{i})/temp)}{{\sum\limits_{j = 0}^{M} \exp((\log(\boldsymbol{\pi}^j_{t1})+z_{j})/temp) }} ,
\end{equation}

where $z_{i}$ is a random number sampled from the Gumbel distribution \cite{gumbel1954statistical} and $temp$ is the temperature of Gumbel-Softmax. In this paper we use a superscript to index the element in a vector.

To generate the $\tau$-th API at time step $t$ when $\tau$ is greater than 1, the decoder RNN uses Formula \ref{equ:generate2} to update the hidden state.

\begin{equation}
\label{equ:generate2}
\boldsymbol{h}^D_{\tau}=Dec(\boldsymbol{W}_{g}\boldsymbol{g}_{t(\tau-1)},\boldsymbol{h}^D_{\tau-1}) .
\end{equation}

The decoder RNN takes the previous Gumbel-Softmax output as input. $\boldsymbol{W}_{g}$ is used to map $\boldsymbol{g}_{t(\tau-1)}$ to a space with the same dimension as $\boldsymbol{h}_t$, in order to make the input dimension of the decoder RNN compatible with Formula \ref{equ:generate1}.

Calculating Gumbel-Softmax for $\tau>1$ can use the same way as $\tau=1$ (i.e. Formula \ref{equ:generate-Gumbel-softmax}). Therefore, we omit the formula here. 

After generating small sequences from $t=1$ to $T$ and inserting the generated sequences to the original sequence, we obtained two kinds of results.

The first kind of result is the one-hot representation of the final adversarial sequence $S_{adv}$:

\begin{equation}
\label{equ:adv}
S_{adv}=RemoveNull(\boldsymbol{x}_1,\boldsymbol{a}_{11},\boldsymbol{a}_{12},...,\boldsymbol{a}_{1L},\boldsymbol{x}_2,\boldsymbol{a}_{21},\boldsymbol{a}_{22},...,\boldsymbol{a}_{2L},......,\boldsymbol{x}_T,\boldsymbol{a}_{T1},\boldsymbol{a}_{T2},...,\boldsymbol{a}_{TL}) .
\end{equation}

The generated null APIs should be removed from the one-hot sequence.

The second kind of result uses Gumbel-Softmax outputs to replace one-hot representations:

\begin{equation}
\label{equ:Gumbel-sequence}
S_{Gumbel}=\boldsymbol{x}_1,\boldsymbol{g}_{11},\boldsymbol{g}_{12},...,\boldsymbol{g}_{1L},\boldsymbol{x}_2,\boldsymbol{g}_{21},\boldsymbol{g}_{22},...,\boldsymbol{g}_{2L},......,\boldsymbol{x}_T,\boldsymbol{g}_{T1},\boldsymbol{g}_{T2},...,\boldsymbol{g}_{TL} .
\end{equation}

The null APIs' Gumbel-Softmax outputs are reserved in the sequence, in order to connect the gradients of loss function with null APIs. The loss function will be defined in the following sections.

\subsection{The Substitute RNN}
Malware authors usually do not know the detailed structure of the victim RNN. They do not know whether the victim RNN uses bidirectional connection, average pooling and the attention mechanism. The weights of the victim RNN is also unavailable to malware authors.

To fit such victim RNN with unknown structure and weights, a neural network with strong representation ability should be used. Therefore, the substitute RNN uses bidirectional RNN with attention mechanism since it is able to learn complex sequential patterns. Bidirectional connection contains both the forward connection and the backward connection, and therefore it has the ability to represent the unidirectional connection. The attention mechanism is able to focus on different positions of the sequence. Therefore, RNN with attention mechanism can represent the cases without attention mechanism such as average pooling and the using of the last state to represent the sequence.  

To fit the victim RNN, the substitute RNN should regard the output labels of the victim RNN on the training data as the target labels. The training data should contain both malware and benign programs.

As shown in Figure \ref{fig:gan} and the previous section, for malware input two kinds of outputs are generated from the generative RNN, i.e. the one-hot adversarial example $S_{adv}$ and the Gumbel-Softmax output $S_{Gumbel}$.

We use the victim RNN to detect the one-hot adversarial example, and get the resulting label $v$. $v$ is a binary value where $0$ represents benign label and $1$ represents malware.

Then the substitute RNN is used to classify the Gumbel-Softmax output $S_{Gumbel}$, and outputs the malicious probability $p_S$.

Cross entropy is used as the loss function, as shown in Formula \ref{equ:substitute-ce}.

\begin{equation}
\label{equ:substitute-ce}
L_S=-v\log(p_S)-(1-v)log(1-p_S) .
\end{equation}

\begin{figure*}[htp]
    \begin{center}
    \graphicspath{{img/}}
    \includegraphics[width = 4.0in]{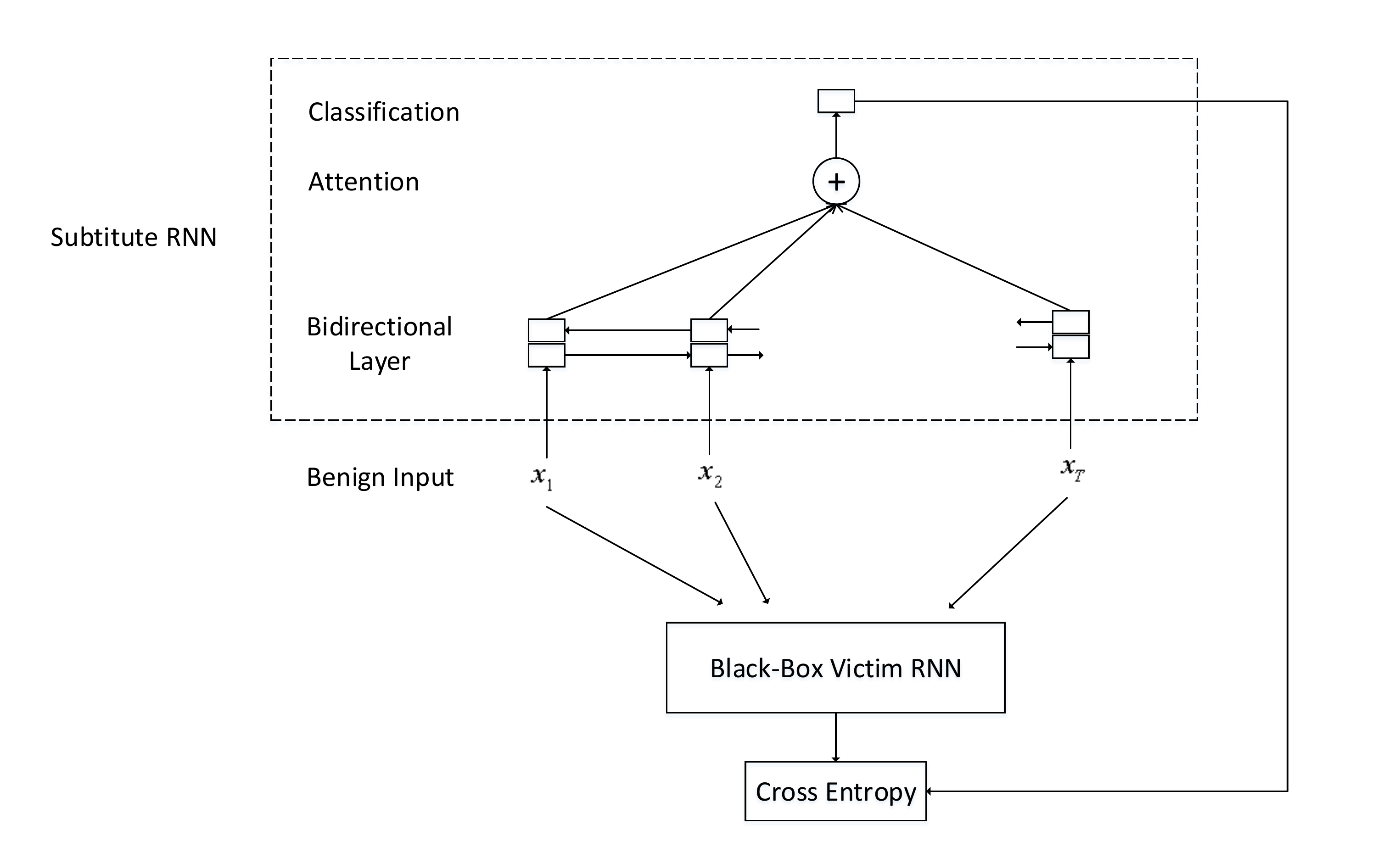}
    \caption{The architecture of the proposed model when trained on benign programs.}
    \label{fig:benign}
    \end{center}
\end{figure*}

For a benign input sequence, it is directly fed into the victim RNN and the substitute RNN, as shown in Figure \ref{fig:benign}. The outputs of the two RNNs $v$ and $p_S$ are used to calculate the loss function in the same way as Formula \ref{equ:substitute-ce}.

\subsection{Training}
The training objective of the generative RNN is to minimize the predicted malicious probability $p_S$ on $S_{Gumbel}$. We also add a regularization term to restrict the number of inserted APIs in the adversarial sequence by maximizing the null API's expectation probability. The final loss function of the generative RNN is defined in Formula \ref{equ:g-loss}.

\begin{equation}
\label{equ:g-loss}
L_G=\log(p_S)-\gamma \mathbb{E}_{t=1\sim T,\tau=1\sim L}\boldsymbol{\pi}^M_{t\tau} ,
\end{equation}

where $\gamma$ is the regularization coefficient and $M$ is the index of the null API.

The training process of the proposed model is summarized in Algorithm \ref{alg:training}.

\begin{algorithm}
\caption{Training the Proposed Model}
\label{alg:training}
\begin{algorithmic}[1]
\While{terminal condition not satisfied}
\State Sample a minibatch of data, which contains malware and benign programs.
\State Calculate the outputs of the generative RNN for malware.
\State Get the outputs of the substitute RNN on benign programs and the Gumbel-Softmax output of malware.
\State Get the outputs of the victim RNN on the adversarial examples and benign programs.
\State Minimize $L_S$ on both benign and malware data by updating the substitute RNN's weights.
\State Minimize $L_G$ on malware data by updating the generative RNN's weights.
\EndWhile

\end{algorithmic}
\end{algorithm}
\section{Experiments}
Adam \cite{kingma2014adam} was used to train all of the models. LSTM unit was used for all of the RNNs presented in the experiments due to its good performance in processing long sequences \cite{hochreiter1997long,greff2016lstm}.

\subsection{Dataset}
We crawled 180 programs with corresponding behavior reports from a website for malware analysis (https://malwr.com/). On the website users can upload their programs and the website will execute the programs in virtual machines. Then the API sequences called by the uploaded programs will be posted on the website. 70\% of the crawled programs are malware.

In real-world applications, the adversarial example generation model and the victim RNN should be trained by malware authors and anti-virus vendors respectively. The datasets that they collected cannot be the same. Therefore, we use different training sets for the two models. We selected 30\% of our dataset as the training set of the adversarial example generation model (i.e. the generative RNN and the substitute RNN), and selected 10\% as the validation set of the adversarial example generation model. Then we selected another 30\% and 10\% as the training set and the validation set of the victim RNN respectively. The remaining 20\% of our dataset was regarded as the test set.





\subsection{The Victim RNNs}
To validate the representation ability of the substitute RNN, we used the several different structures for the black-box victim RNN, as shown in the first column of Table \ref{tab:victim-results}. In Table \ref{tab:victim-results}, the first LSTM model uses the last hidden state as the representation of the sequence. BiLSTM represents bidirectional LSTM. The suffixes ``Average'' and ``Attention'' in the last four rows indicate the use of average pooling and attention mechanism to represent the sequence.


We first tuned the hyper-parameters of BiLSTM-Attention on the validation set. The final learning rate was set to 0.001. The number of recurrent hidden layers and the number of attention hidden layers were both set to one and the layer sizes were both set to 128. We directly used the resulting hyper-parameters to other victim models. We have tried to separately tune the hyper-parameters for other victim RNNs but the performance did not improve much compared with using BiLSTM-Attention's hyper-parameters.

Table \ref{tab:victim-results} gives the area under curve (AUC) of these victim RNNs before adversarial attacks.

\begin{table}[htbp]
  \centering
  \caption{AUC of different victim RNNs before attacks.}
    \begin{tabular}{lll}
    \toprule
    Algorhthm & Training Set & Test Set  \\
          \midrule
    LSTM    & 94.57\% & 91.30\% \\
    BiLSTM    & 94.67\% & 92.80\% \\
    LSTM-Average   & 93.07\% & 92.66\% \\
    BiLSTM-Average  & 91.13\% & 91.14\% \\
    LSTM-Attention   & 95.98\% & 93.97\% \\
    BiLSTM-Attention   & 95.02\% & 93.83\% \\
    \bottomrule
    \end{tabular}%
  \label{tab:victim-results}%
\end{table}%

Overall, the attention mechanism works better than non-attention approaches, since attention mechanism is able to learn the relative importance of different parts in sequences.
LSTM and BiLSTM only use the last hidden state, and therefore the information delivered to the output layer is limited. In this case bidirectional connection delivers more information than unidirectional connection, and AUC of BiLSTM is higher than LSTM.
For average pooling and attention mechanism, bidirectional LSTM does not outperform unidirectional LSTM in AUC.
Average pooling and attention mechanism are able to capture the information of the whole API sequence. Unidirectional LSTM is enough to learn the sequential patterns. Compared with unidirectional LSTM, bidirectional LSTM has more parameters, which makes the learn process more difficult. Therefore, the bidirectional connection does not improve the performance for average pooling and attention mechanism.

\subsection{Experimental Results of the Proposed Model}

The hyper-parameters of the generative RNN and the substitute RNN were tuned separately for each black-box victim RNN. The learning rate and the regularization coefficient were chosen by line search along the direction 0.01, 0.001, et al.. The Gumbel-Softmax temperature was searched in the range $[1, 100]$. Actually, the decoder length $L$ in the generative RNN is also a kind of regularization coefficient. A large $L$ will make the generative RNN have strong representation ability, but the whole adversarial sequences will become too long, and the generative RNN's size may exceed the capacity of the GPU memory. Therefore, in our experiments we set $L$ to 1.

The experimental results are show in Table \ref{tab:results}.

\begin{table}[htbp]
  \centering
  \caption{Detection rate on original samples and adversarial examples. ``Adver." represents adversarial examples.}
  \resizebox{0.7\linewidth}{!}{
    \begin{tabular}{llrllr}
    \toprule
    & \multicolumn{2}{l}{Training Set} & & \multicolumn{2}{l}{Test Set}  \\
    \cline{2-3}\cline{5-6}
          & Original& Adver.& & Original & Adver. \\
          \midrule
    LSTM    & 92.54\% & 12.10\% & & 90.74\% & 11.95\% \\
    BiLSTM    & 92.21\% & 1.06\% & & 90.93\% & 0.95\% \\
    LSTM-Average    & 93.87\% & 1.40\% & & 93.53\% & 1.36\% \\
    BiLSTM-Average   & 92.92\% & 1.83\% & & 92.51\% & 1.67\% \\
    LSTM-Attention    & 93.67\% & 0.44\% & & 92.45\% & 0.51\% \\
    BiLSTM-Attention   & 93.73\% & 3.02\% & & 92.99\% & 3.03\% \\
    \bottomrule
    \end{tabular}}%
  \label{tab:results}%
\end{table}%

After adversarial attacks, all the victim RNNs fails to detect most of the malware. For different victim RNNs, the detection rates on adversarial examples range from 0.44\% to 12.10\%, while before adversarial attacks the detection rates range from 90.74\% to 93.87\%. That is to say, about 90\% malware will bypass the detection algorithms under our proposed attack model.

Except LSTM, the detection rates on adversarial examples of all the victim models are smaller than 3.03\%, which means that victim RNNs are almost unable to detect any malware. The victim model LSTM have detection rates of 12.10\% and 11.95\% on the training set and the test set respectively, which are higher than other victim RNNs. We can see that for the LSTM model the substitute RNN does not fit the victim RNN very well on the training data.

The differences in adversarial examples' detection rates are very small between the training set and the test set for these victim RNNs. The generalization ability of the proposed model is quite well for unseen malware examples. The proposed adversarial example generation algorithm can be applied to both existing malware and unseen malware.

It can be seen that even if the adversarial example generation algorithm and the victim RNN use different RNN models and different training set, most of the adversarial examples are still able to attack the victim RNN successfully. The adversarial examples can transfer among different models and different training sets. The transferability makes it very easy for malware authors to attack RNN based malware detection algorithms.

\section{Conclusions and Future Works}
A novel algorithm of generating sequential adversarial examples for malware is proposed in this paper. The generative network is based on the sequence to sequence model. A substitute RNN is trained to fit the black-box victim RNN. We use Gumbel-Softmax to approximate the generated discrete APIs, which is able to propagate the gradients from the substitute RNN to the generative RNN. The proposed model has successfully made most of the generated adversarial examples able to bypass several black-box victim RNNs with different structures.

Previous researches on adversarial examples mainly focused on images which have fixed input dimension. We have shown that the sequential machine models are also not safe under adversarial attacks. The problem of adversarial examples becomes more serious when it comes to malware detection. Robust defensive models are needed to deal with adversarial attacks.

In future works we will use the proposed model to attack convolutional neural network (CNN) based malware detection algorithms, since many researchers have begun to use CNN to process sequential data recently \cite{zhang2015character,lee2016sequential}. We will validate whether a substitute RNN has enough capacity to fit a victim CNN, and whether a substitute CNN has enough capacity to fit a victim RNN. The research on the transferability of adversarial examples between RNN and CNN is very important to the practicability of sequential malware detection algorithms.

\bibliographystyle{plain}
\bibliography{references}

\end{document}